\documentclass[runningheads]{llncs}

\usepackage[T1]{fontenc}
\usepackage[misc,geometry]{ifsym}
\usepackage[marginal]{footmisc}

\def\doi#1{\href{https://doi.org/\detokenize{#1}}{\url{https://doi.org/\detokenize{#1}}}}
\usepackage{graphicx}
\usepackage{amsmath}
\usepackage{amsfonts}
\usepackage{bbm}
\usepackage{listings}
\usepackage{xcolor, soul}

\begin{document}
\title{Coarse Retinal Lesion Annotations Refinement via Prototypical Learning}
\author{Qinji Yu\inst{1} \and Kang Dang\inst{2} \and Ziyu Zhou\inst{1}\and Yongwei Chen\inst{2}\and Xiaowei Ding\inst{1,2}\textsuperscript{(\Letter)}}
\institute{Shanghai Jiao Tong University, Shanghai, China  \and VoxelCloud, Inc., Los Angeles, USA}
\authorrunning{Q. Yu et al.}
\maketitle              

\renewcommand{\thefootnote}{}
\footnote{Q. Yu and K. Dang contribute equally to this work.}

\begin{abstract}
Deep-learning-based approaches for retinal lesion segmentation often require an abundant amount of precise pixel-wise annotated data. However, coarse annotations such as circles or ellipses for outlining the lesion area can be six times more efficient than pixel-level annotation. Therefore, this paper proposes an annotation refinement network to convert a coarse annotation into a pixel-level segmentation mask. Our main novelty is the application of the prototype learning paradigm to enhance the generalization ability across different datasets or types of lesions. We also introduce a prototype weighing module to handle challenging cases where the lesion is overly small. The proposed method was trained on the publicly available IDRiD dataset and then generalized to the public DDR and our real-world private datasets. Experiments show that our approach substantially improved the initial coarse mask and outperformed the non-prototypical baseline by a large margin. Moreover, we demonstrate the usefulness of the prototype weighing module in both cross-dataset and cross-class settings.
\keywords{Prototypical Learning  \and Retina Lesion Segmentation \and Coarse Annotation Refinement.}
\end{abstract}

\section{Introduction}
Given the growing demand for retinal screening, automatic segmentation for retinal lesions enjoys increasing clinical relevance. By answering the issue of what lesions exist in the image and where they are located, retinal lesion segmentation algorithms assist ophthalmologists in making clinical diagnoses and assessing disease severity~\cite{wei2021learn}. While recent deep-learning approaches have tremendously boosted the retinal lesion segmentation accuracy~\cite{wei2021learn,liu2021m2mrf,huang2022rtnet,yan2019learning}, they often require abundant expert-level-accurate, pixel-wise annotated data, which requires significant time and expense to acquire. Previous studies show that coarse annotations such as circles or ellipses for outlining the lesion area can be six times more efficient than pixel-level annotation~\cite{huang2020automated}. Therefore, it is essential to study novel methodologies tailored for lower-quality coarse annotations.

Existing works on exploiting coarse annotations can be categorized into weakly-supervised segmentation~\cite{playout2019novel,yang2018boxnet,tang2021weakly,liu2022weakly,wang2021bounding,chu2021improving,zhang2021refinemask}, and mask refinement~\cite{yang2019label,huang2020automated}. Weakly-supervised segmentation methods rely on prior assumptions such as box tightness constraint~\cite{wang2021bounding} and image contrast constraint~\cite{tang2021weakly} to utilize box-level and image-level coarse annotations. A few high-quality pixel-level retinal lesion datasets such as {\cal{IDRiD}}~\cite{porwal2020idrid} and {\cal{DDR}}~\cite{li2019diagnostic} provide precise lesion boundaries. While successful, weakly-supervised segmentation does not utilize these pre-existing lesion segmentation datasets that provide rich knowledge on lesions' exact appearance and shape. Instead of further developing weakly-supervised segmentation methods, we propose to use such datasets by training an annotation refinement model in a data-driven manner to convert a coarse annotation into a pixel-level segmentation mask. It should be emphasized that our work is significantly different from the weakly-supervised approach, as it is trained in a fully supervised way (instead of weakly-supervised) with coarse annotation and pixel-level ground truth in pairs. Additionally, we note several existing mask refinement methods~\cite{yang2019label,huang2020automated} which refine initial coarse masks into more accurate segmentation results; however, they are usually optimized for a particular dataset. In comparison, our method applies the prototype learning paradigm~\cite{wang2019panet,li2021adaptive,tang2021recurrent,zhang2021refinemask} to enhance generalization across different data sets and lesion types. Good generalization is the key to putting the coarse annotation refinement algorithm into practice. For example, we can train the coarse annotation refinement network on a large-scale dataset for once and reuse the trained model on other datasets with less fine-tuning or tweakings.

Particularly, our prototype learning averages the features from the coarse mask region to form a lesion prototype and averages the background features to create a background prototype. A pixel is classified to the lesion class if its corresponding feature vector is more similar to the lesion prototype. Since our prototypical approach generates image-specific prototype to adaptively describe the image itself, it is less sensitive to the intra-class variance and high distribution shifts from different datasets or unseen classes. However, averaging features uniformly may be problematic when the lesion is considerably smaller than the coarse mask, as the resultant lesion prototype  becomes dominated by background features. We alleviate this issue by a superpixel-guided prototype weighing module. The module first divides the coarse mask into several superpixels\cite{li2021adaptive} and the prototype for each superpixel is obtained. Each prototype's dis-similarity with the background prototype is then calculated as a weighting factor. The final lesion prototype is the weighted combination of these superpixel-guided sub-prototypes.

\textbf{Contributions.} (1) To the best of our knowledge, our method is the first prototypical approach for the coarse retinal lesion annotation refinement problem. (2) We present a prototype weighing module to solve the problem of the actual lesion being overly small. (3) Experiments demonstrate that the proposed method substantially improved the initial coarse annotation and outperformed non-prototypical mask refinement baselines. It also confirms the superiority of the prototype weighing module in both cross-dataset and cross-class settings.

\section{Methods}
\begin{figure}[htb]
	\centering{\includegraphics[width=300pt]{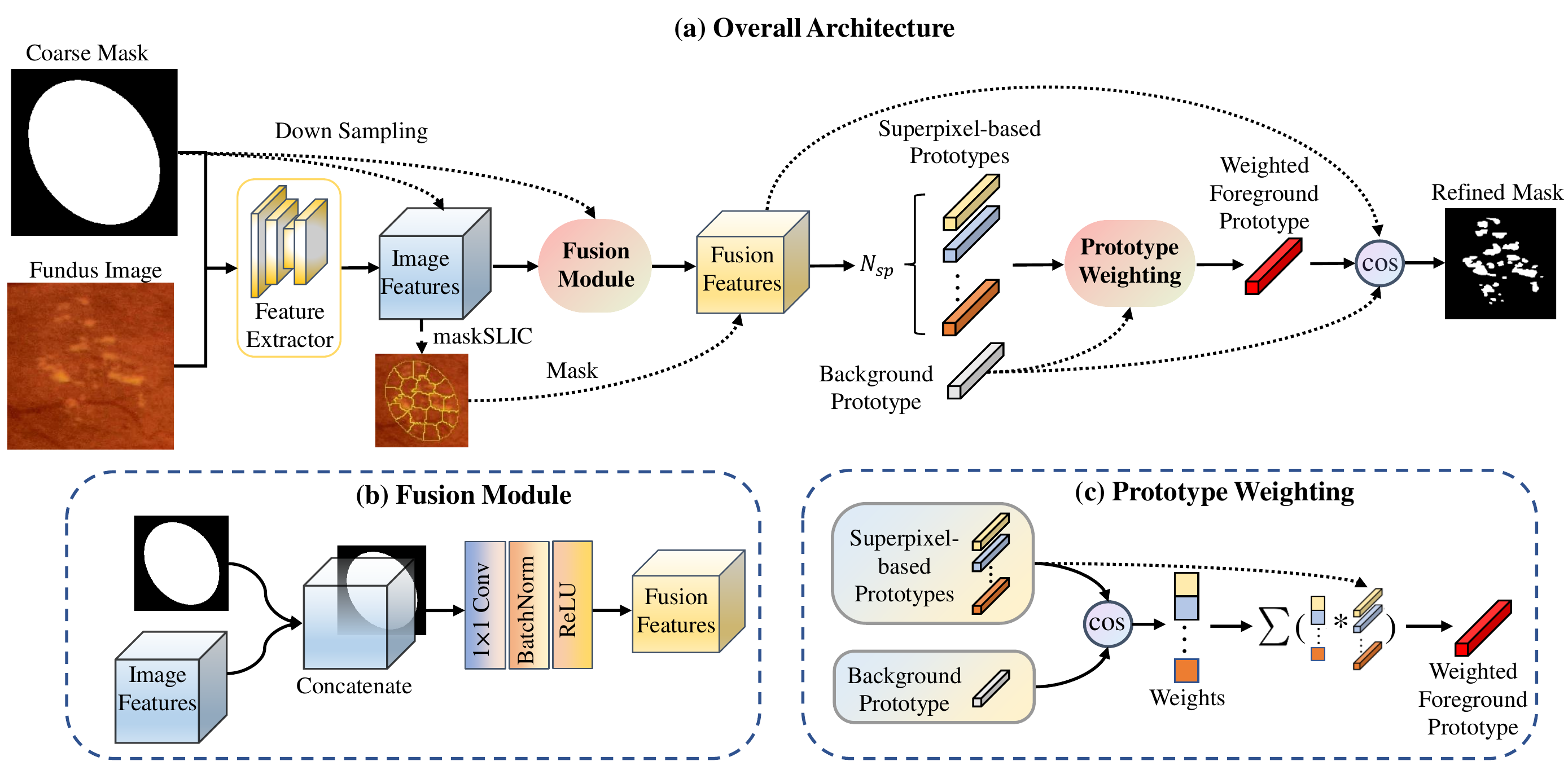}}
	\caption{Framework of our prototype-based coarse annotation refinement network.} \label{fig1}
\end{figure}

This section details the proposed coarse annotation refinement method with the overall structure shown in Fig. \ref{fig1}.  We assume that there exists a set of image patches and the associated coarse lesion annotations, and our algorithm will convert them into the corresponding high-quality pixel-level annotations.

\subsection{Annotation Refinment via Prototype Learning} 
\subsubsection{Feature extraction \label{feature_extraction}}  The input to the network is the concatenation of the image patch $I\in\mathbb{R}^{H\times W\times3}$ and its corresponding coarse lesion annotation $M\in\mathbb{R}^{H\times W\times 1}$. We use a modified U-Net backbone to extract its feature map $F\in\mathbb{R}^{H^\prime\times W^\prime\times C}$. Following \cite{tang2021recurrent}, we remove the last two upsampling blocks in the U-Net to speed up the calculation. As a result, the resolution of the feature map is 1/4 of the original input. We concatenate the feature map with the down-sampled coarse mask $M'\in\mathbb{R}^{H'\times W'\times 1}$ in the feature channel dimension to further incorporate the coarse annotation prior. To get the final fusion feature map $F^\prime\in\mathbb{R}^{H^\prime\times W^\prime\times C^\prime}$, we adopt a simple 1-layer network with architecture: 1$\times$1 Conv2d$+$BatchNorm2d$+$ReLU.

\subsubsection{Coarse Prototype Extraction} Given the fused feature map, we want to learn representative and well-separated prototype vectors for the lesion region and the background based on the prototypical network.  In previous research \cite{wang2019panet,tang2021recurrent,yu2021location}, the prototypical network condenses the masked object features in an image into a single or few prototypes. A relative simple coarse foreground lesion prototype can be calculated by mask average pooling, as follows:
\begin{equation}
	p_{fg}=\frac{\sum_{(x,y)} F'(x,y)\mathbbm{1}[M'(x,y)=1]}{\sum_{(x,y)}\mathbbm{1}[M'(x,y)=1]},
\end{equation}
where $(x,y)$ indexes the spatial locations and $\mathbbm{1}(\bullet)$ is an indicator function. In addition, the background prototype is computed by
\begin{equation}
	p_{bg}=\frac{\sum_{(x,y)} F'(x,y)\mathbbm{1}[M'(x,y)=0]}{\sum_{(x,y)}\mathbbm{1}[M'(x,y)=0]},
\end{equation} 
where $p_{fg}, p_{bg}\in\mathbb{R}^{C'}$.

\subsubsection{Coarse Annotation Refinement} Refinement is done using a non-parametric metric learning method \cite{wang2019panet}. For each pixel at location $(x,y)$ of the final fusion feature map $F'$, we calculate the distance between its feature vector and the derived prototypes $\mathcal{P} =\left\{p_{bg}{,p}_{fg}\right\}$. Then, we apply the softmax operation over the distances to get the probability map $P_c\in\mathbb{R}^{H'\times W'\times1}$ and $c \in \{bg,fg \}$. Formally, we have:
\begin{equation}
	P_c(x,y)=\frac{\text{exp}(-\alpha \cdot d(F'(x,y),p_c))}{\sum_{p_j \in \mathcal{P}}\text{exp}(-\alpha \cdot d(F'(x,y),p_j))},
\end{equation}
where $\alpha$ is the scaling factor fixed at 20. 

We train our model end-to-end using the sum of dice loss $\mathcal{L}_{dice}$ and binary cross-entropy loss $\mathcal{L}_{bce}$ between the final probability map $P_{fg}$ and the well-annotated ground truth mask $M_{gt}$. That is: $\mathcal{L}_{loss} = \mathcal{L}_{dice} + \mathcal{L}_{bce}$.  

During testing, for each image patch $I$ and its corresponding coarse lesion annotation $M$, we obtain a corresponding foreground probability map $P_{fg}$. When mapping $P_{fg}$ back to the original image space (uncropped full image), some of them will overlap. For each pixel in the overlapping area, we choose the maximum probability of these probability maps as its value. In the end, thresholding is used to get the final refined mask.

\begin{figure}
	\centering{\includegraphics[width=300pt]{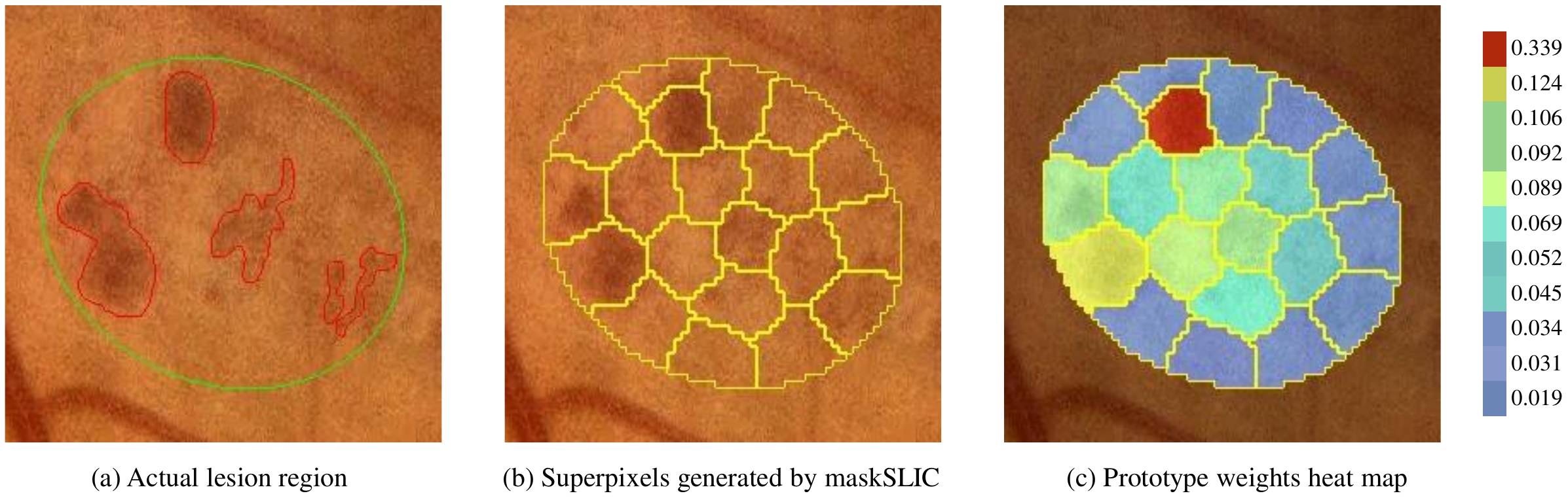}}
	\caption{Illustration of superpixel-guided prototype weighting. Boundaries in red, green, and yellow are actual lesion, coarse lesion, and superpixel regions respectively} \label{fig2}
\end{figure}

\subsection{Superpixel-guided prototype weighting } As shown in Fig. \ref{fig2} when the actual lesion is relatively small compared to the coarse mask, the coarse foreground prototype defined by Eq. (1) cannot represent the actual lesion features. To reduce the impact of false-positive pixels within the coarse annotation, we divide the initial coarse region into several sub-regions according to their feature similarity. Concretely, we refer to maskSLIC \cite{irving2016maskslic} to aggregate the feature map within the masked region into multiple superpixel clusters. For each superpixel region $S_i$, we can obtain its corresponding superpixel-based sub-prototype $g_{i}$ according to Eq. (1). We collect the extracted sub-prototypes and denote them as set $\mathcal{G}=\{ g_{i}\} $, where $i\in{1,2,\cdots,N_{sp}}$ ($N_{sp}$ is the number of superpixels). We compute the cosine distance to measure the similarity between each $g_{i}$ and $p_{bg}$: 
\begin{equation}
	d(g_{i},p_{bg})=1-\frac{g_{i} \cdot  p_{bg}}{\left\| g_i \right\| \cdot \left\| p_{bg} \right\| }.
\end{equation}

Intuitively, the prototypes dissimilar to the background prototype are more important parts for the final foreground prototype. Therefore, we can derive a weight coefficient for each prototype in set $\mathcal{G}$:
\begin{equation}
	w_{i}=\frac{\text{exp}(\beta \cdot d(g_{i},p_{bg}))}{\sum_{g_j \in \mathcal{G}}\text{exp}(\beta \cdot d(g_{j},p_{bg}))}, 
\end{equation}
where $\beta$ is the scaling factor fixed at 10. The final foreground prototype is then given by
\begin{equation}
	p_{\text{weighted}}\ =\ \sum_{g_{i}\in\mathcal{G}}w_{i}\cdot g_{i}.
\end{equation}
As shown later, our proposed superpixel weighted prototype $p_{\text{weighted}}$ is a more representative foreground prototype that performed better in various experiments.

\section{Experiments and Results}
\subsection{Experimental Setup}
\subsubsection{Coarse Annotation Generation} There is no public available retinal lesion dataset with paired coarse annotation and pixel-level segmentation mask. To construct such paired dataset, we develop a simple coarse annotation generation method. Firstly, the coarse annotations are simulated from the well-annotated fine masks by applying the following chain of operations: smoothing, dilating, expanding, clustering the connected components using DBSCAN \cite{ester1996density} and fitting ellipses to each cluster. Secondly, the fundus image is cropped around each ellipse in the corresponding coarse annotation. Finally, these cropped image patch and coarse annotation pairs are resized to fixed dimensions $H\times W$ for subsequent model training and testing.

\subsubsection{Datasets and Evaluation Metrics.} We evaluate the proposed methods on publicly IDRiD and DDR datasets, and our real-world private dataset. IDRiD contains 81 fundus images (54 training images, 27 testing images) with pixel-level annotations for hard exudates (EX), hemorrhages (HE), microaneurysms (MA), and soft exudates (SE). Similarly, the testing part of DDR contains 225 fundus images with pixel-level annotations for EX, HE, MA, and SE. Our real-world private testing dataset collects 211 fundus images with pixel-level annotations for drusen (Drus) and pre-retinal hemorrhages (Prh) labeled by two experienced ophthalmologists. To train our annotation refinement network, we collect 32985 training patch pairs (EX:9957, HE:7752, MA:14387, SE:889) from the IDRiD training images using our coarse annotation generation algorithm. We also apply the mask generation algorithm to generate coarse mask for each testing image. To compare different refinement methods, we calculate the Intersection over Union (IoU) between the refined annotation and ground-truth mask.

\subsubsection{Baselines.} We implement several non-prototypical mask refinement baseline models, taking in an image patch and a coarse mask as the input. In detail, we choose three widely used feature extraction backbones, Res18 \cite{he2016deep}, HRNet18 \cite{YuanCW19},  and U-Net \cite{ronneberger2015u} attached with the coarse mask fusion module to perform feature extraction. The feature extraction process is identical to the one described in Sec.\ref{feature_extraction} except for the feature backbone. After that, we attach a 1x1 Conv2d layer as a binary classifier to obtain a refined segmentation score map.

\subsubsection{Implementation Details.} For the prototypical methods, we set the superpixel number $N_{sp} = 20$. All training patch pairs are resized to 256 $\times$ 256 and augmented by RandomShiftScaleRotate, RandomBlur, and RandomBrightnessContrast. All models are implemented by PyTorch and trained from scratch using Adam optimizer with a batch size of 64 for 120 epochs. The initial learning rate is $10^{-4}$ and reduces according to ReduceLROnPlateau strategy.

\subsection{Results}
\subsubsection{Same-Dataset Experiments}  We train the proposed coarse annotation refinement network using all four lesion types on IDRiD and evaluate the performance on the IDRiD testing set. As shown in Table \ref{tab1}, “Initial Coarse” denotes
the IoU of actual lesion region versus coarse annotation region. Our prototypical method improves the initial coarse mask considerably. It also consistently obtains better refinement performance than the non-prototypical baselines on all four lesion classes in terms of IoU score, with average IoU score improving by more than 5.2\%.  This experiment demonstrates our advantages when training and testing images are from the same dataset.

\begin{table}
\centering
\setlength{\tabcolsep}{0.3mm}
\caption{The image-level average IoU (\%) and its standard deviations (\%) of on IDRiD. "w/ superpixel" means with superpixel-guided prototype weighing.}\label{tab1}
\begin{tabular}{l|cccc|c}
\hline
Methods & MA & SE & EX & HE & Average\\
\hline
Initial Coarse & 9.6 (2.6) & 49.3 (10.4) & 15.0 (5.2) & 33.2 (8.6)& 26.8 (6.7)\\
\hline
Res18 \cite{he2016deep} & 73.9 (7.3) & 68.4 (14.2) & 54.1 (9.4) & 62.6 (10.9)& 64.7 (10.5)\\
HRNet18 \cite{YuanCW19} & 79.1 (7.5) & 78.1 (5.3) & 56.8 (9.1) & 64.8 (11.8)& 69.7 (8.4)\\
U-Net \cite{ronneberger2015u} & 77.6 (6.8) & 75.6 (12.6) & 58.9 (8.9) & 67.9 (11.4)& 70.0 (9.9)\\
\hline
Our methods & \textbf{84.2} (6.2)  & \textbf{80.7} (9.2) & 65.3 (8.1) & 69.9 (13.9) & 75.0 (9.4)\\
w/ superpixel & 84.1 (6.2) & 79.6 (9.9) & \textbf{65.9} (8.7) & \textbf{71.1} (11.2) & \textbf{75.2} (9.0)\\
\hline
\end{tabular}
\end{table}

\subsubsection{Cross-Dataset and Cross-Class Experiments} We directly evaluate the performance on the DDR testing set and our real-world private dataset using models trained on the IDRiD dataset without further fine-tuning. As shown in Table \ref{tab2}, our method exceeds the U-Net baseline by 4.3\% on both DDR and private datasets. For DDR, our superpixel weighted prototype performs better for all lesion types compared to the non-weighted prototype. Similarly, the weighted prototype is notably better than the non-weighted one on the private dataset, especially for the class Prh (56.3\% $\rightarrow$ 62.6\%). Overall, we see a general trend that our model can generalize well to new datasets or unseen classes.

\begin{table}
    \centering
    \caption{The image-level average IoU (\%) and its standard deviations (\%) on DDR and our real-world private dataset having 2 unseen classes.}\label{tab2}
    \resizebox{\linewidth}{!}{
    \begin{tabular}{l|ccccc|ccc}
    \hline
      & & & DDR  & & & & Private  &\\
    \hline
    Methods & MA & SE & EX & HE & Average & Drus & Prh & Average\\
    \hline
    Initial Coarse & 6.9 (3.4) & 32.3 (10.8) & 14.8 (8.8) & 23.1 (11.4) & 19.3 (8.6) &33.6 (19.1) &49.7 (13.0) & 41.6 (16.0)\\
    \hline
    Res18 \cite{he2016deep} & 56.3 (12.4) & 67.6 (13.6)  & 52.3 (13.2) & 54.4 (13.3) & 57.6 (13.1) &43.9 (16.6) &52.9 (15.2)  &48.4 (15.9)\\
    HRNet18 \cite{YuanCW19} & 65.1 (13.5) & 68.2 (12.6) & 54.2 (13.9) & 58.5 (12.9)& 61.5 (13.3) &44.2 (21.1) &57.5 (16.4) &50.9 (18.8)\\
    U-Net \cite{ronneberger2015u} & 60.9 (13.2) & 70.2 (15.0) & 55.4 (12.8) & 59.8 (12.7) & 61.6 (13.4) &45.2 (20.9)&57.9 (16.4) &51.6 (18.6)\\
    \hline
    Our methods & 68.8 (12.0) & 71.2 (16.9) & 58.4 (12.7) & 61.3 (15.7)& 64.9 (14.3) &47.9 (23.4)  &56.3 (20.7) &52.1 (22.1)\\
    w/ superpixel & \textbf{69.8} (11.9) & \textbf{72.0} (17.5) & \textbf{58.9} (12.4) & \textbf{62.9} (14.5) & \textbf{65.9} (14.1) &\textbf{49.1} (23.0) &\textbf{62.6} (15.3) &\textbf{55.9} (19.1)\\
    \hline
    
    \end{tabular}}
    \end{table}

\subsubsection{Coarse Mask Reduction Factors} 
\begin{figure}[htb]
    \centering{\includegraphics[width=280pt]{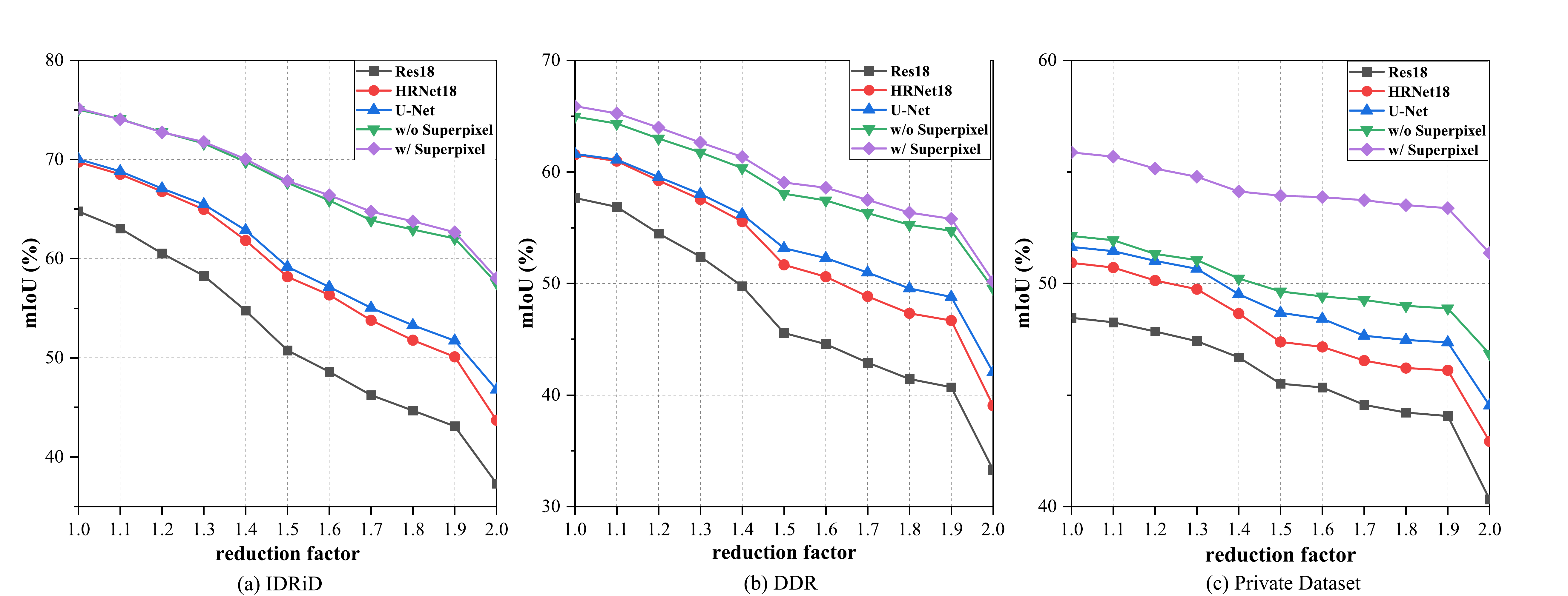}}
    \caption{Refinement performance under different reduction factors.} \label{fig3}
    \end{figure}

Since ophthalmologists tend to draw a single rough ellipse to cover several unconnected lesion regions, we simulate the process by setting different reduction factors to the DBSCAN clustering algorithm. Actually, the number of the generated ellipses is the number of connected lesion regions divided by the reduction factor. In other words, with a higher reduction factor, the generated coarse mask will be more coarse.  As shown in Fig. \ref{fig3}, although the refinement performance of all methods degrades as the reduction factor ranges from 1.0 to 2.0, our prototypical method has less degradation compared to the U-Net baseline. It implies our method is more robust against coarser annotations.

\subsubsection{Visual results} 
\begin{figure}[htb]
    \centering{\includegraphics[width=300pt]{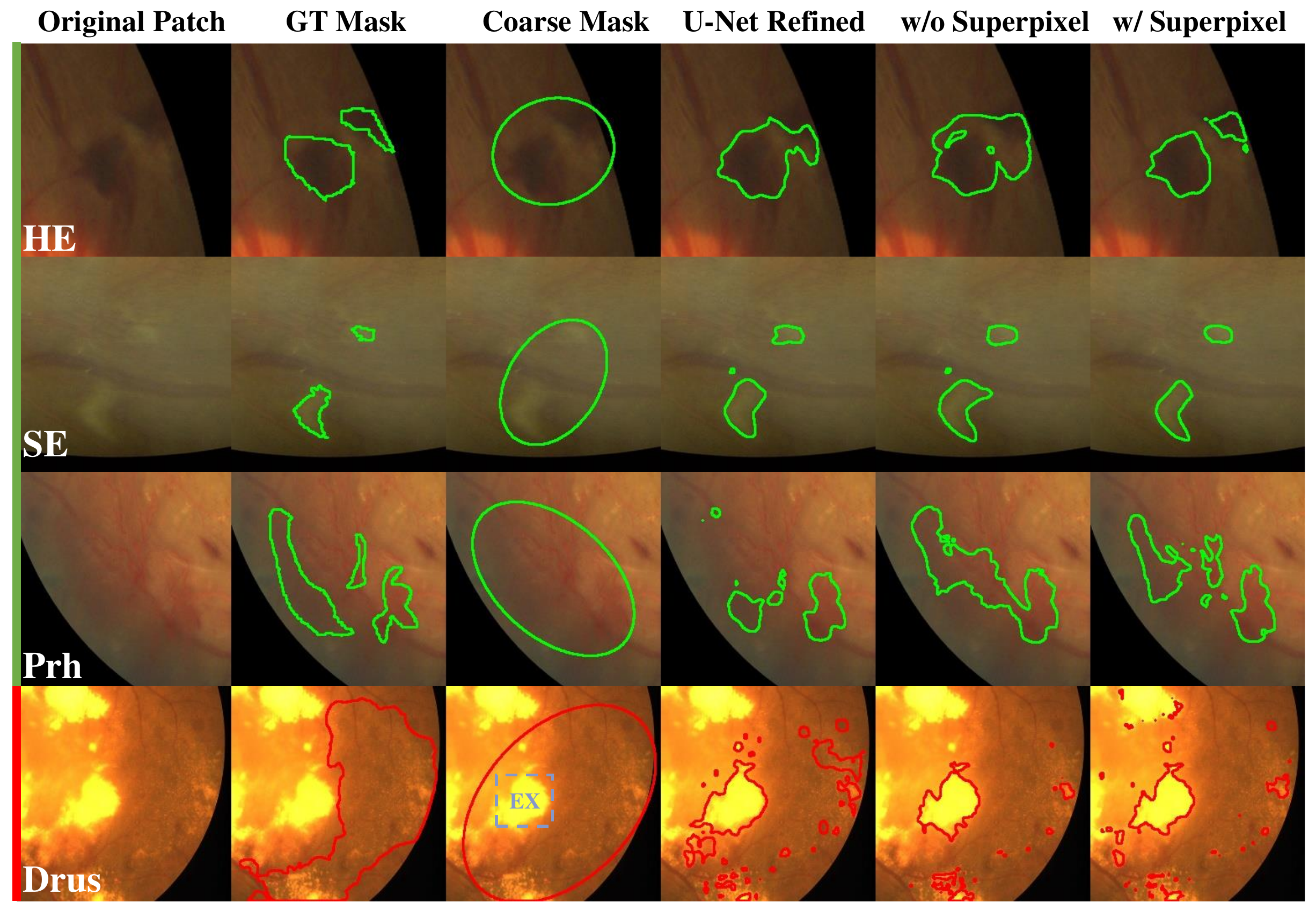}}
    \caption{Visualization of refinement results on four categories of lesions.} \label{fig4}
    \end{figure}

Fig. \ref{fig4} presents some visualization of refinement results. Despite the vast variation in lesion scales, colors, and low contrast to surrounding regions, the first three rows show our proposed superpixel weighted prototype approach generates the most accurate lesion boundary. The last row shows a failure case when the coarse mask contains two distinct lesion classes, EX and Drus, at the same time. 

\section{Conclusions}
This paper proposes a novel prototype-based network to convert a coarse annotation into a pixel-level segmentation mask. The proposed network first extracts the lesion and background prototypes and labels the image pixel as the lesion class if its feature is more similar to the lesion prototype. A superpixel-guided prototype weighing module is then proposed to tackle the issue of the actual lesion being overly small compared to the coarse mask. On the IDRiD dataset, our model outperformed non-prototypical baselines by a large margin. Extensive experiments on DDR and our real-world private dataset also demonstrate the proposed model enjoys better generalizability to new datasets and some unseen lesion classes.

%
%
%
\bibliographystyle{splncs04}
\bibliography{refs}

\end{document}